# An Enhancement of CNN Algorithm for Rice Leaf Disease Image Classification in Mobile Applications

Kayne Uriel K. Rodrigo[1], Jerriane Hillary Heart S. Marcial[2], Samuel C. Brillo[3], Khatalyn E. Mata[4], Jonathan C. Morano[5]

[1]*Student in Computer Science, Computer Science Department, College of Information Systems and Technology Management, University of the City of Manila, Manila, Philippines.*
[2]*Student in Computer Science, Computer Science Department, College of Information Systems and Technology Management, University of the City of Manila, Manila, Philippines.*
[3]*Faculty in Biology, Biology Department, College of Science, University of the City of Manila, Manila, Philippines.*
[4]*Dean in Computer Science, Computer Science Department, College of Information Systems and Technology Management, University of the City of Manila, Manila, Philippines.*
[5]*Faculty in Computer Science, Computer Science Department, College of Information Systems and Technology Management, University of the City of Manila, Manila, Philippines.*

*Email: [1]kukrodrigo2021@plm.edu.ph, [2]jhhsmarcial2021@plm.edu.ph, [3]scbrillo@plm.edu.ph, [4]kemata@plm.edu.ph, [5]jcmorano@plm.edu.ph*
*Orcid Id number: [1]0009-0004-8953-3686, [2]0009-0006-5300-6299, [3]0009-0008-7018-2985, [4]0000-0003-1413-6120, [5]0009-0007-1353-334X*
*Corresponding Author\*: Kayne Uriel K. Rodrigo*

| ARTICLE INFO | ABSTRACT |
|---|---|
| Received: 12 Oct 2024<br><br>Revised: 09 Dec 2024<br><br>Accepted: 26 Dec 2024 | This study focuses on enhancing rice leaf disease image classification algorithms, which have traditionally relied on convolutional neural network (CNN) models. We employed transfer learning with MobileViTV2_050 using ImageNet-1k weights—a lightweight model that integrates CNN's local feature extraction with Vision Transformers' global context learning through a separable self-attention mechanism. Our approach resulted in a significant 15.66% improvement in classification accuracy for MobileViTV2_050-A, our first enhanced model trained on the baseline dataset, achieving 93.14%. Furthermore, MobileViTV2_050-B, our second enhanced model trained on a broader rice leaf dataset, demonstrated a 22.12% improvement, reaching 99.6% test accuracy. Additionally, MobileViTV2-A attained an F1-score of 93% across four rice labels and a Receiver Operating Characteristic (ROC) curve ranging from 87% to 97%. In terms of resource consumption, our enhanced models reduced the total parameters of the baseline CNN model up to 92.50%, from 14 million to 1.1 million. These results indicate that MobileViTV2_050 not only improves computational efficiency through its separable self-attention mechanism but also enhances global context learning. Consequently, it offers a lightweight and robust solution suitable for mobile deployment, advancing the interpretability and practicality of models in precision agriculture.<br><br>**Keywords:** Rice Leaf Disease Image Classification, Transfer Learning, MobileViTV2, Convolutional Neural Networks (CNN), Precision Agriculture, Model Efficiency. |

## INTRODUCTION:

Rice serves as a staple food for many people worldwide, making the cultivation of rice remains critical. In the Philippines, millions of Filipino farmers rely on rice as the primary source of income, and it also supports environmental sustainability [1]. In the 2022/2023 period, global rice consumption has reached 520.4 million metric tons, underscoring its importance and the increasing reliance on rice due to population growth [2]. Ensuring rice production security is challenging due to environmental factors, specifically the spread of diseases observable through leaf quality. These diseases typically appear during the growth phase of the plant, especially the heading stage, leading to significant loss in rice crop yields [3]. Fortunately, advancements in artificial intelligence (AI), particularly in machine learning algorithms that are suitable for mobile devices, enable high-quality inputs to perform human-like tasks. Rice leaf disease detection leverages AI using images processed through algorithmic techniques. Several studies indicate that convolutional neural networks (CNN) are the preferred choice for analyzing plant leaf diseases, including rice leaves [4]. Despite the advantages of convolutional neural networks (CNNs), achieving optimized models for rice leaf disease classification remains challenging. Recent systematic reviews identified critical gaps, highlighting the need for robust, computationally efficient architectures, lightweight models with fewer parameters, and the implementation of transfer learning to overcome the limitations of training deep learning methods from scratch. Additionally, there was a demand for models that offer explainable diagnoses to address the "black box" nature of CNNs, enhancing the interpretability of predictions [5]. Researchers explored attention mechanisms to improve model interpretability and efficiency, with one study utilizing an attention-based neural network and Bayesian optimization to achieve a test accuracy of 94.65%, outperforming traditional CNN architectures such as VGG16 and ResNet50 [6]. Other studies employed hybrid CNN and Vision Transformer (ViT) models to classify five primary rice diseases, though these models were manually constructed and limited in scope [7]. To address these limitations, newer models like MobileViT were developed, combining CNN and ViT architectures to provide global information processing suitable for mobile devices and demonstrating superior performance with fewer data points [8, 9, 10]. The plant-based MobileViT (PMVT) further enhanced disease classification by focusing on significant features, though it suffered from slower inference speeds and higher memory requirements, limiting its deployment on resource-constrained mobile devices [11]. In response, the separable self-attention method led to MobileViTV2, which reduced time complexity from quadratic to linear, enhancing processing speed and memory efficiency for mobile deployment [12]. However, MobileViTV2 had not yet been applied to rice leaf disease classification, presenting a field for exploration. In this context, researchers utilized a CNN model with five convolutional and two dense layers, where it has a 78% accuracy. It encounters overfitting and classification challenges with diverse textures using datasets from Kaggle [13]. Utilizing the Timm library in [14], the researchers trained the MobileViTV2 model under the MobileViTV2_050 lightweight variant using the baseline dataset indicated in [13] for baseline comparison and the enhanced dataset from [15] for general comparison. By doing so, the model leverages its separable self-attention mechanism to efficiently learn global image contexts while maintaining lower memory consumption and time complexity, thereby enabling effective deployment on mobile edge devices through validated evaluation metrics.

## METHODOLOGY:

A standardized hybrid deep learning approach was applied to enhance the traditional rice leaf disease image classification. It integrates the local feature extraction from the Convolutional Neural Network (CNN) with the global context learning capabilities of the Vision Transformer (ViT). A lightweight MobileViTV2 model was evaluated against conventional CNN models to determine its efficiency and accuracy.

### a. Research Design

An experimental research design was employed to enhance a standard convolutional neural network (CNN) from [13] by integrating the MobileViTV2 architecture. This addresses the novel application of MobileViTV2 in rice leaf disease classification for testing. Table 1 summarizes the comparative analysis of model components, highlighting the enhanced model's superior classification performance and optimized resource usage, thereby demonstrating its suitability for deployment in mobile applications for plant disease detection. Hence, the study involved three training sessions: first, the baseline CNN from [13] for simulating the methods of the study; the MobileViTV2 trained with a baseline dataset called `mobilevitv2_050-A` for directly assessing the performance improvements with identical data conditions. The second is a MobileViTV2 model with enhanced dataset, labeled as `mobilevitv2_050-B`, wherein it was retrained using a broader Kaggle-sourced dataset verified by a university biologist to evaluate generalizability across diverse rice leaf disease images. Comparative analyses were conducted in two parts: firstly, comparing the baseline CNN against the mobilevitv2_050-A model using the same validation metrics. Secondly, comparing the performance of mobilevitv2_050-B with reported comparative metrics from several benchmark models from other studies. Furthermore, all benchmark models utilize attention mechanisms in rice leaf disease classification, acknowledging that benchmark models were not retrained with similar datasets due to time and resource constraints. The evaluation encompassed classification metrics (accuracy, precision, recall, F1-score) and efficiency metrics (Top 1, Top 5, total parameters, FLOPs, FPS), providing a comprehensive assessment of both performance and computational efficiency.

**Table 1.** Comparative Analysis of Models, from BaseLinkapd

| Attribute | Baseline CNN (5 Conv + 2 Dense) | MobileViTV2 (mobilevitv2_050 A & B) |
|---|---|---|
| *Total Parameters* | 14 M | 1.1 M |
| *Model Size (MB)* | **56.20 MB** | **4.26 MB** |
| *Input Shape* | (None, 256, 256, 3) | [Batch Size, 3, 224, 224] |
| *Output Classes* | 4 Classes | 4 Classes (A), 10 Classes (B) |
| *Layer Information* | 5 Convolutional Layers | Convolutional Layers Integrated Within Blocks |
| | 2 Dense Layers | MobileViTV2 Blocks (Multiple) |
| | 4 Batch Normalization Layers | Layer Transformer Blocks (Multiple) |
| | 3 Dropout Layers | BatchNormAct2d Layers (Multiple) |
| *Activation Functions* | ReLU | SiLU |
| *Attention Mechanisms* | None | Linear Self-Attention within MobileViTV2 Blocks |

b. **Materials and Equipment**

The study utilized the following hardware and software resources during training and evaluation time, respectively: an IdeapadGaming3 laptop equipped with AMD Ryzen 5 5600H with Radeon Graphics and 24 GB RAM. B.) Conda environment with Python 3.11.9 and the PyTorch deep learning framework. The MobileViTV2 model was sourced from the Timm library, a Pytorch Image Model library known for computer vision pre-trained models.

c. **Data Acquisition and Cleaning**

The enhanced MobileViTV2 model utilized two datasets. First, from [13], to have the same dataset in comparison with the baseline CNN model. Second, from [15] which is a broader rice leaf disease dataset from Kaggle, to investigate the generalizability of the model across several rice leaf disease image datasets. In the first dataset from [13], it consists of 4 classes specified in Table 2. In the second dataset from [15], it consists of 18,445 rice leaf disease images, encompassing ten distinct categories including healthy leaves and various paddy disease types, as shown in Table 2. The images were sourced from Kaggle, a public data science repository. The proposed dataset was validated by a university biologist following Philippine Rice Research Institute (PhilRice) guidelines to ensure accurate disease label classification. During the data cleaning phase, nomenclature corrections were implemented, such as renaming the 'neck blast' folder to 'panicle blast' to address misleading image categorization.

**Table 2.** Dataset Summary

| Leaf Classes (Baseline) | No. of Images | Training set size (80%) | Validation set size (20%) | Percentage |
|---|---|---|---|---|
| *Brown Spot* | 523 | 418 | 105 | 15.59% |
| *Healthy* | 1,488 | 1,190 | 298 | 44.36% |
| *Hispa (disease)* | 565 | 452 | 113 | 16.84% |
| *Leaf Blast* | 779 | 623 | 156 | 23.22% |
| **Total** | **3,355 images** | **2683 images** | **672** | **100%** |
| **Leaf Classes (Proposed)** | **No. of Images** | **Training set size (80%)** | **Validation set size (20%)** | **Percentage** |
| *Bacterial Leaf Blight* | 1,762 | 1,410 | 352 | 9.55% |
| *Brown Spot* | 1,860 | 1,488 | 372 | 10.08% |
| *Healthy* | 1,882 | 1,506 | 376 | 10.20% |
| *Leaf Blast* | 2,163 | 1,730 | 433 | 11.73% |
| *Leaf Scald* | 2,056 | 1,645 | 411 | 11.15% |
| *Narrow Brown Leaf Spot* | 1,798 | 1,438 | 360 | 9.75% |
| *Panicle Blast* | 1,322 | 1,058 | 264 | 7.17% |
| *Rice Hispa Disease* | 1,686 | 1,349 | 337 | 9.14% |
| *Sheath Blight* | 1,866 | 1,493 | 373 | 10.12% |
| *Tungro* | 2,050 | 1,640 | 410 | 11.11% |
| **Total** | **18,445 images** | **14,756 images** | **3,689 images** | **100%** |

#### d. Model Architecture: MobileViTV2 Enhancement
- **Overview of MobileViTV2**

Figure 1 illustrates the MobileViTV2 architecture, and the block utilized in this study, designed for efficient rice leaf disease classification. The model processes input images with dimensions of 3 channels × 224 × 224 pixels. The initial convolutional layer (Conv2d) reduces the spatial dimensions to 16 channels × 112 × 112 pixels, followed by a Sigmoid Linear Unit (SiLU) activation and batch normalization. The backbone comprises alternating Bottleneck Block and MobileViTV2Block layers, which preserve local feature integrity while incrementally expanding channel dimensions from 16 to 256 and reducing spatial dimensions through a hierarchical sequence (112 → 56 → 28 → 14 → 7/8 pixels). This structure facilitates multiscale feature extraction essential for capturing complex patterns. Central to the backbone are the MobileViTV2 Transformer Blocks, which integrate depth-wise separable convolutions and point-wise convolutions to efficiently process feature maps. Each Transformer Block unfolds the spatial dimensions into vector representations, which are then processed by a separable self-attention mechanism. Unlike traditional self-attention that relies on dot-product operations, MobileViTV2 employs element-wise multiplication to maintain computational efficiency. The attention mechanism is embedded within a feed-forward network and incorporates skip connections to ensure stable gradient flow and enhance model robustness.

Throughout the network, additional linear self-attention mechanisms from Linear Transformer Blocks modules capture long-range dependencies without significantly increasing computational overhead. Dropout layers and identity residual connections are incorporated to further stabilize training and prevent overfitting. The final classification head consists of an adaptive average pulling layer that reduces spatial dimensions to 1×1 pixel, followed by flattening into a 256-dimensional vector and a dense layer mapping to the ten rice leaf disease classes. The mobilevitv2_050 variant features 1.1 million trainable parameters and occupies 4.26 MB of memory, excluding the requirements for forward and backward passes. This lightweight and efficient design renders MobileViTV2 highly suitable for deployment on mobile edge devices, ensuring robust feature extraction with minimal resource consumption.

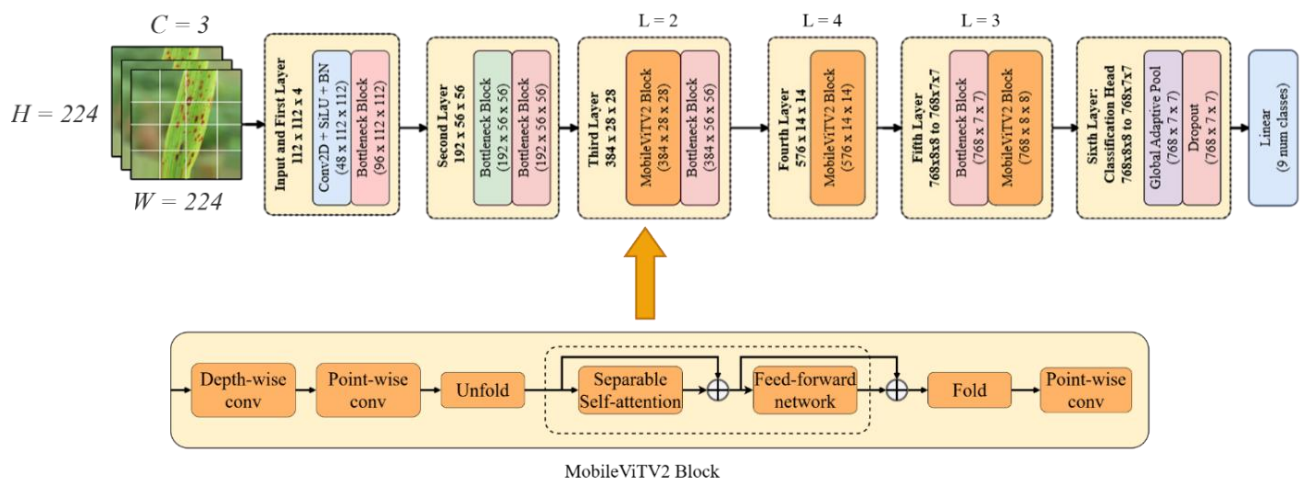

**Figure 1:** MobileViTV2 Model Architecture with MobileViTV2 Block diagram

- **Incorporation of Separable Self-Attention Mechanism**

Figure 2 shows the separable-self attention mechanism. This serves as the underlying attention mechanism within the MobileViTV2 block from [12]. According to the study, it was inspired by Multi-Head Attention from Transformer Encoder Block. However, the enhancement offers a more efficient approach when computing attention scores by reducing the complexity from quadratic $O(k^2)$ to linear $O(k)$, where k represents the number of input tokens. The process begins with three parallel paths: input (I), where it converts each token into a simplified representation; key (K), where it creates keys for cross-check matching of information; and value (V), where it prepares values for information transfer. Secondly, it underwent a latent node approach, wherein instead of comparing the tokens from every path to each other, it shall be redirected to a single reference point called "latent node" (L). This offers a centralized coordinate, reducing its computational complexity. Thirdly, as a result, it will generate a context score (cs), which contains context information about the token, which will then be passed to a Soft Max activation function to reduce their scale. Fourth, the MobileViTV2 transformer block combines all key information from context scores into a context vector (cv), which captures the important features from the entire sequence. Lastly, the transformed versions of input ($xV$) are combined with the context vector. A final transformation produces the output with Rectified Linear Unit (ReLU) added to add non-linearity. Overall, the purpose of these processes was to make attention mechanisms efficient for longer sequences. These are present within the transformer block of MobileViTV2.

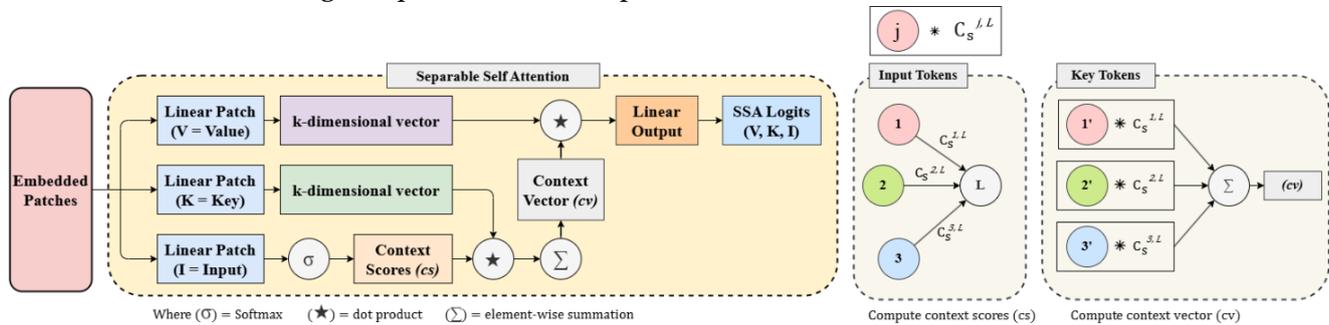

**Figure 2:** Separable Self-Attention Mechanism Diagram from [12]

- **Linear Transformer Block Modules**

This transformer block implements a linear self-attention mechanism by instead using Soft Max-based attention, where it uses pairwise interactions across tokens $Q(K^t)$, the Linear Transformer employed a kernel-based formulation $Q(K^T V)$, reducing the time complexity from $O(n^2)$ to $O(n)$. While the self-attention mechanism has been changed, the feed-forward network remains unchanged, giving n linear transformations and a ReLU activation function.

- **Bottleneck Block**

The bottleneck residual block aims to reduce computational complexity while preserving performance through a three-layer design. The block used a 1 × 1 convolution to decrease the number of channels to 1, creating a bottleneck, followed by a 3 × 3 convolution using a lower dimensional space. Then it applied 1 × 1 convolution to restore the original dimensions. Alongside this, it applied a skip connection to serve as a residual path, where information is retained despite reducing its complexity. It simply aims to have stable training regardless of the network depth.

e. **Experimental Procedures**

The researchers implemented the model using the Timm library for computer vision, specifically leveraging the MobileViTv2_050 variant with 1.4 million parameters to ensure lightweight support for edge devices. The researchers configured the model via timm.create_model(), initializing it with pretrained ImageNet-1k weights and setting num_classes to 10 to match the rice leaf disease categories. For dataset preprocessing, it follows the following steps: first, they split the data into 80% training and 20% validation; second, they resized all rice images to 224×224 pixels and converted them to PyTorch tensors; lastly, they normalized the pixel values using ImageNet mean and standard deviation. During training, they employed the Adam optimizer with a learning rate of 0.001 and utilized the ReduceLROnPlateau scheduler to adjust the learning rate based on validation loss, alongside an Early Stopping mechanism to terminate training when validation loss ceased to improve. The training of the enhanced model will be in two phases: first, by using the same dataset used by the baseline CNN model to identify direct improvements in comparison with the baseline model; second, by using the new dataset publicly available in Kaggle, where it consists of datasets with broader features and rice labels. The model is trained for a maximum of 10 to 20 epochs with a batch size of 32, completing the process in hours. After training, they saved the MobileViTv2_050 model as a PyTorch .pth file and exported it to the ONNX format to ensure broader compatibility with mobile devices.

f. **Data Analysis and Validation Techniques**

The researchers evaluated the trained model's accuracy using metrics such as accuracy and loss scores, classification reports, Receiver Operating Characteristic (ROC) curves, and confusion matrices. The accuracy score as seen in equation (1) assessed overall predictive correctness, while the loss score in equation (2) measured the discrepancy between true and predicted labels on both training and validation datasets. Classification reports provided precision seen in equation (3), recall seen in equation (4), and F1-scores seen in equation (5) for each rice leaf class, enabling a comprehensive analysis of predictions on unseen data. ROC curves illustrated the model's diagnostic performance across various thresholds, and confusion matrices displayed true positives, false negatives, false positives, and true negatives for each class. Top 1 and Top 5 accuracy, seen in equation (6) and equation (7), were added to evaluate the true class among the first and first 5 predictions. Additionally, the researchers assessed algorithm performance by calculating floating point operations per second (FLOPs), seen in equation (8), and frames per second (FPS) to determine inference speed, and measured memory consumption to ensure efficiency on edge devices. These evaluations collectively determined the model's accuracy and operational performance.

$$Accuracy\ Score = \frac{TP + TN}{TP + TN + FP + FN} \quad (1)$$

$$F1\ Score = 2 \cdot \frac{Precision \cdot Recall}{Precision + Recall} \quad (5)$$

$$Loss\ Score = -\frac{1}{N}\sum_{i=1}^{N}\sum_{j=1}^{C} y_{ij}\log(\hat{y}_{ij}) \quad (2)$$

$$Top\ 1 = \frac{No.\,of\ Correct\ Top\ 1}{Total\ No.\,Predictions} \quad (6)$$

$$Precision = \frac{TP}{TP + FP} \quad (3)$$

$$Top\ 5 = \frac{No.\,of\ Correct\ Top\ 5}{Total\ No.\,Predictions} \quad (7)$$

$$Recall = \frac{TP}{TP + FN} \quad (4)$$

$$FLOPS = \frac{No.\,of\ Floating\ Point\ Ops}{Execution\ Time} \quad (8)$$

g. **Application Platform for Plant Disease Detection**

The application platform used for integrating the model was a Kotlin-based ONNX Runtime application. ONNX Runtime offers a variety of options to integrate machine learning models, including the trained mobilevitv2_050 model to mobile applications, and can measure inference time in real time. The researchers utilized its given mobile image classification template deployed in the Android operating system. Part of its user interface design would include that while the camera is open, it will continuously detect rice leaf disease images and scans for potential rice labels in real time. For testing and deployment, it was tested on images that do not belong to the dataset, such as sample images from International Rice Research Institute documents, and it accurately identifies the disease, given the correct distance used. Figure 3 shows the prototype of the ONNX Runtime application.

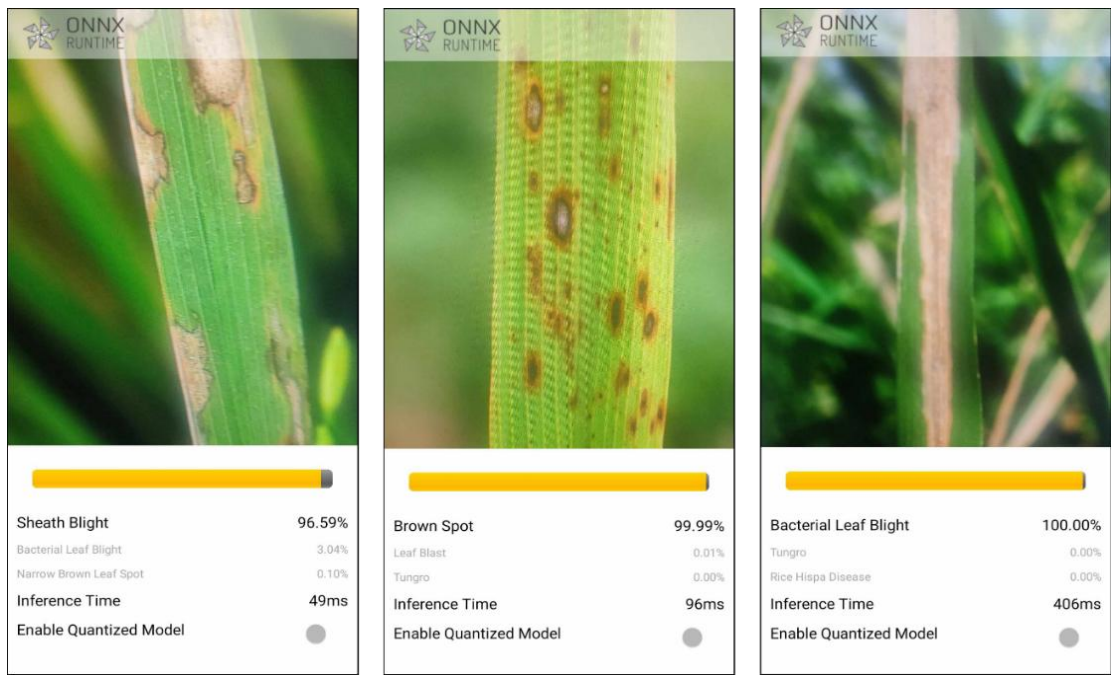

**Figure 3** The MobileViTV2_050 deployed in ONNX Runtime application

## RESULTS AND DISCUSSION

Table 3 shows the results that came from the baseline and the enhanced model. The mobilevitv2_050-A, a variant of the MobileViTV2 model trained with baseline, demonstrates a superior performance across all evaluation metrics when compared to the baseline CNN model. The enhanced model surpasses the validation accuracy of the baseline model by 15.66%, while it also exhibits notable improvements in disease detection capabilities across all rice labels, as it was higher and offers a lower accuracy interval. The enhanced model also shows exceptional precision, recall, and F1-Scores, as it received a consistent output of 93% to 94%. These performance improvements establish MobileViTV2's effectiveness in rice leaf disease image classification, presenting a robust solution that enhances the reliability and accuracy of disease diagnosis in agricultural management and crop protection systems.

**Table 3.** Comparative Analysis of Baseline CNN Model vs. Enhanced Model from MobileViTV2-A

| Existing CNN Model (5-Conv & 2 dense) | | | | |
|---|---|---|---|---|
| **Disease Labels** | **Accuracy** | **Precision** | **Recall** | **F1-Score** |
| *Brown Spot* | 85.42% | 70% | 61% | 65% |
| *Healthy* | 85.1% | 52% | 87% | 65% |
| *Hispa (disease)* | 65.18% | - | - | - |
| *Leaf Blast* | 64.56% | 27% | 12% | 17% |
| **Total Score** | **77.48%** | **41%** | **52%** | **44%** |
| Enhanced Model (via mobilevitv2_050-A) | | | | |
| **Disease Labels** | **Accuracy** | **Precision** | **Recall** | **F1-Score** |
| *Brown Spot* | 98.21% | 93% | 96% | 94% |
| *Healthy* | 94.49% | 90% | 99% | 94% |
| *Hispa (disease)* | 96.42% | 99% | 80% | 88% |
| *Leaf Blast* | 97.17% | 98% | 89% | 93% |
| **Total Score** | **93.14%** | **94%** | **93%** | **93%** |

Alongside the accuracy of performance is the evaluation of resource consumption to identify its compatibility with mobile-edge devices. Table 4 shows a comparative analysis of the baseline CNN model (5-conv & 2 dense CNN), the enhanced model with the baseline dataset (mobilevitv2_050-A), and the enhanced model trained in a broader dataset (mobilevitv2_050-B). The mobilevitv2_050-B variant emerges as the superior performer, achieving exceptional Top 1 accuracy of 99.6% and maintaining almost perfect precision and recall, resulting in an F1-score of 97%. Notably, both enhanced variants demonstrate a substantial efficiency improvement against the baseline CNN model, reducing its parameters down to 13.6 M, leaving 1.1 M, which implies that it enters the threshold of a lightweight model. Additionally, the floating-point operation values were reduced down to 0.92 M, leaning to 0.36 GFLOPs. However, computational efficiency comes with a trade-off in inference speed, as MobileViTV2 variants process fewer images per second compared to the baseline CNN model. This suggests that while MobileViTV2, an optimized hybrid CNN-ViT variant, significantly enhances the accuracy and reduces model complexity, they require additional optimization for real-time applications, where processing speed is critical.

**Table 4.** Comparative Analysis of Baseline CNN Model against Enhanced Model from MobileViTV2

| Method Name | Top 1 Accuracy | Top 5 Accuracy | F1-Score | Precision | Recall | Total No. of Params | FLOPs (G) | FPS (Images per Second) |
|---|---|---|---|---|---|---|---|---|
| *5-conv & 2 dense CNN* | 43.89% | 100% | 44% | 54% | 56% | 14.7 M | 1.28 | **111.33** |
| *mobilevitv2_050-A* | 93.14% | 100% | 93% | 94% | 93% | 1.1 M | 0.36 | 49.20 |
| *mobilevitv2_050-B* | **99.6%** | 100% | 97% | **99.6%** | **99.6%** | **1.1 M** | **0.36** | 44.6 |

Figure 4 presents the visual comparison of model performance between the baseline CNN and enhanced MobileViTV2_050-A model models through ROC curves and confusion matrices. In Figure 4 A and B, the ROC curves reveal a significant improvement in classification performance. The enhanced model in (B) demonstrates near-perfect curves with rapid accession towards the left corner, indicating superior true positive rates across all disease classes, which contrasts in (A), which shows a lower gradual slope and lower area under the curve (AUC) values. The confusion matrices in panels C and D simply show that the mobilevitv2_050-A in D indicates a clear visualization of classification accuracy due to a consistent diagonal coloring. Here, most true positives lie in the healthy label due to darker contrast, which was caused by the dataset itself. On the other hand, C shows a dispersed coloring, indicating lower classification accuracy and higher rates of misclassification of rice labels. Hence, it implies that mobilevitv2_050-A under MobileViTV2 represents superior discriminative and performance capabilities.

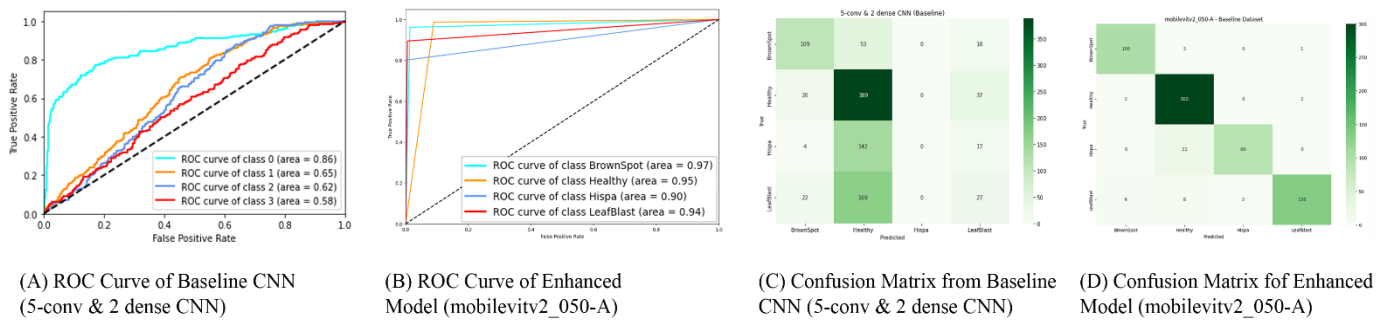

(A) ROC Curve of Baseline CNN (5-conv & 2 dense CNN)    (B) ROC Curve of Enhanced Model (mobilevitv2_050-A)    (C) Confusion Matrix from Baseline CNN (5-conv & 2 dense CNN)    (D) Confusion Matrix fof Enhanced Model (mobilevitv2_050-A)

**Figure 4:** Comparison of Multi-class Rice Leaf Disease Detection Performance. (A) ROC Curves of Baseline CNN; (B) ROC Curves of Enhanced MobileViTV2_050-A; (C) Confusion Matrix of Baseline CNN; (D) Confusion Matrix of Enhanced MobileViTV2_050-A

- **Comparison of MobileViTV2_050-B vs. Other Backbone Models**

To further enhance the generalizability of the proposed model, a comparative analysis of related studies was used to compare distinctive results in lightweight model development with integrated attention mechanisms. Figure 5 shows the ROC curve and confusion matrix of mobilevitv2_050-B. It implies that the mobilevitv2 architecture under the enhanced dataset, containing 10 labels, provides remarkable accuracy, given the highly pointed ROC curve towards the upper left and a consistent diagonal lining in the confusion matrix. Table 4 demonstrates that the mobilevitv2_050-B achieves optimal performance with minimal computational overhead, delivering 99.6% top 1 accuracy while maintaining the lowest floating-point operations. The mobilevitv2_050-B variant, with its reduced parameter count from 14M to 1.1M, establishes itself as a lightweight CNN-ViT hybrid architecture that balances computational efficiency with high accuracy performance.

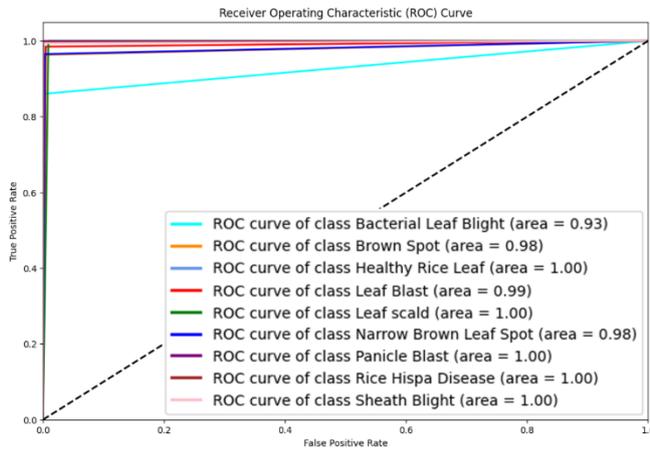
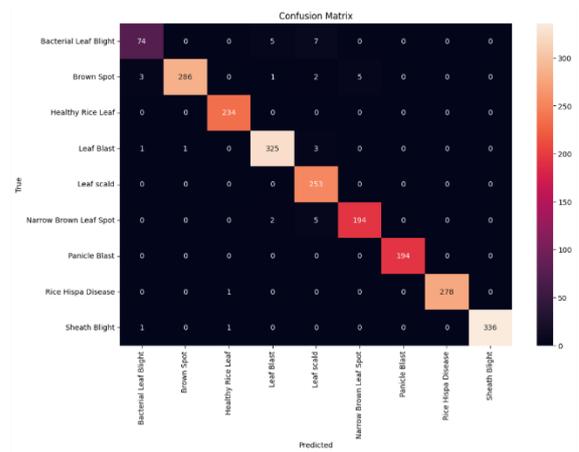

(A) ROC Curves of Enhanced MobileViTV2_050-B          (B) Confusion Matrix of Enhanced MobileViTV2_050-B

**Figure 5:** Visualization of accuracy results from the MobileViTV2_050-B variant

**Table 4.** Comparative Analysis of MobileViTV2_050-B vs. Other Benchmark Studies

| Method Name | Top 1 Accuracy (Rice) | Total Parameters | FLOPs (G) | FPS (Images per second) |
|---|---|---|---|---|
| ViT-CNN Hybrid | 100.0% | - | - | - |
| mobilevitv2_050-B (ours) | **99.6%** | **1.1 M** | **0.36** | 44.6 |
| PMVT-XS | 97.7% | 2.01 M | 0.85 | 85.3 |
| ADSNN-B0 | 94.7% | - | - | - |
| PMVT-XXS | 93.1% | 0.98 M | 0.31 | 88.5 |
| PMVT-S | 92.0% | 5.06 M | 1.59 | 81.3 |

## CONCLUSION:

This study presented an enhanced approach to baseline CNN-based rice leaf disease classification by integrating MobileViTV2. This novel method combines CNN's feature extraction capabilities with global context learning from Vision Transformers and introduces a separable self-attention mechanism, reducing computational complexity. The enhanced model demonstrated significant improvements in accuracy and resource efficiency compared to the baseline CNN, which is widely used in existing studies. By addressing key research gaps such as the CNN model's "black box" nature, limited global context understanding, and high computational demands. MobileViTV2 not only improves classification performance but also optimizes efficiency for resource-constrained devices. This results in a lightweight and robust solution for rice leaf disease image classification in precision agriculture. The implications of these findings are substantial for precision agriculture. The MobileViTV2 model's computational efficiency makes it highly suitable for deployment on various mobile edge devices, enabling farmers, including those with visual impairments or limited experience, to accurately classify rice leaf diseases in the field. This advancement can lead to increased rice production and reduce unnecessary disease treatments.

## ACKNOWLEDGEMENT

We thank Kayne Uriel Rodrigo, Jerriane Hillary Heart Marcial, and Samuel Brillo for training and validating the datasets, testing the model, algorithm, and architecture of this study. Special thanks to Dr. Khatalyn Mata and Prof. Jonathan Morano for their assistance and understanding throughout the study.

## REFERENCES:

[1] C. B. Cororaton, J. Cockburn, and E. Corong, "Doha Scenarios, Trade Reforms, and Poverty in the Philippines A CGE Analysis," SSRN Electronic Journal, vol. 57, 2005, doi: https://doi.org/10.2139/ssrn.774346.
[2] Shahbandeh M, "Total Global Rice Consumption 2019 | Statista," Statista, 2019. https://www.statista.com/statistics/255977/total-global-rice-consumption/

[3] Philippine Rice Research Institute "Learning Modules - Pinoy Rice Knowledge Bank," Pinoy Rice Knowledge Bank, 2014. https://www.pinoyrice.com/resources/learning-modules/.

[4] Wubetu Barud Demilie, "Plant disease detection and classification techniques: a comparative study of the performances," Journal of Big Data, vol. 11, no. 1, Jan. 2024, doi: https://doi.org/10.1186/s40537-023-00863-9.

[5] H. M. Yusuf, S. A. Yusuf, A. H. Abubakar, M. Abdullahi, and I. H. Hassan, "A systematic review of deep learning techniques for rice disease recognition: Current trends and future directions," Franklin Open, vol. 8, p. 100154, Sep. 2024, doi: https://doi.org/10.1016/j.fraope.2024.100154.

[6] Y. Wang, H. Wang, and Z. Peng, "Rice diseases detection and classification using attention based neural network and bayesian optimization," Expert Systems with Applications, vol. 178, p. 114770, Sep. 2021, doi: https://doi.org/10.1016/j.eswa.2021.114770.

[7] Tharani Pavithra P and Baranidharan B, "A Hybrid ViT-CNN Model Premeditated for Rice Leaf Disease Identification," International Journal of Computational Methods and Experimental Measurements, vol. 12, no. 1, pp. 35–43, Mar. 2024, doi: https://doi.org/10.18280/ijcmem.120104.

[8] A. Dosovitskiy et al., "An Image is Worth 16x16 Words: Transformers for Image Recognition at Scale," arXiv:2010.11929 [cs], Oct. 2020. https://arxiv.org/abs/2010.11929

[9] A. Vaswani et al., "Attention Is All You Need," arXiv, Jun. 12, 2017. https://arxiv.org/abs/1706.03762.

[10] S. Mehta and M. Rastegari, "MobileViT: Light-weight, General-purpose, and Mobile-friendly Vision Transformer," arXiv.org, 2021. https://arxiv.org/abs/2110.02178.

[11] G. Li et al., "PMVT: a lightweight vision transformer for plant disease identification on mobile devices," 2023, doi: https://doi.org/10.3389/fpls.2023.1256773.

[12] S. Mehta and M. Rastegari, "Separable Self-attention for Mobile Vision Transformers," arXiv.org, 2022. https://arxiv.org/abs/2206.02680.

[13] P. Tejaswini, P. Singh, M. Ramchandani, Y. K. Rathore, and R. R. Janghel, "Rice Leaf Disease Classification Using Cnn," IOP Conference Series: Earth and Environmental Science, vol. 1032, no. 1, p. 012017, Jun. 2022, doi: https://doi.org/10.1088/1755-1315/1032/1/012017.

[14] Lokesh R, "Rice Leaf Diseases Detection," Kaggle.com, 2023. https://www.kaggle.com/datasets/loki4514/rice-leaf-diseases-detection/data

[15] Hugging Face, "timm/mobilevitv2_050.cvnets_in1k · Hugging Face," *Huggingface.co*, Jul. 16, 2024. https://huggingface.co/timm/mobilevitv2_050.cvnets_in1k (accessed Nov. 09, 2024).